\documentclass[11pt]{article}
\usepackage{geometry}
\usepackage{coling2020}
\usepackage{times}
\usepackage{url}
\usepackage{latexsym}
\usepackage{microtype}
\usepackage{booktabs}
\usepackage{graphicx}
\usepackage{mathtools}
\usepackage{url}

\colingfinalcopy %

\title{defx at SemEval-2020 Task 6:  Joint Extraction of Concepts and Relations for Definition Extraction}
        
\author{Marc Hübner, Christoph Alt, Robert Schwarzenberg, Leonhard Hennig \\ German Research Center for Artificial Intelligence (DFKI), Berlin, Germany \\ \texttt{\{firstname.lastname@dfki.de\}}}

\date{}

\begin{document}
\maketitle
\begin{abstract}
Definition Extraction systems are a valuable knowledge source for both humans and algorithms. In this paper we describe our submissions to the DeftEval shared task (SemEval-2020 Task 6), which is evaluated on an English textbook corpus. We provide a detailed explanation of our system for the joint extraction of definition concepts and the relations among them. Furthermore we provide an ablation study of our model variations and describe the results of an error analysis.
\end{abstract}

\section{Introduction}

\blfootnote{
    This work is licensed under a Creative Commons Attribution 4.0 International License. License details: \url{http://creativecommons.org/licenses/by/4.0/}.
}

Definition extraction (DE) is a subfield of information extraction that deals with the automated extraction of terms and their descriptions from natural language text. We belief that for humans definitions are one of the most important sources of knowledge to clarify unknown words in a new language or domain.
Thus, an important application is the automated construction of a glossary, which is a laborious task if done manually. Another important application is the use of definitions as background knowledge for machine learning algorithms. Recent results have disclosed difficulties of state of the art approaches when confronted with factual knowledge~\cite{logan-etal-2019-baracks}, and have lead to a surge of research in this area~\cite{logeswaran-etal-2019-zero,ernie,knowbert}. The advantages of definitions as a knowledge source are two-fold:~Firstly, definitions are easier to extract and annotate than other forms of knowledge sources, e.g.\ Knowledge Graphs (KG). And secondly, definitions can be processed using the same text-based algorithms whereas the use of a KG requires other forms of algorithms, e.g., Graph Embeddings.
While research using definitions exists, it is limited by the scale of available annotations and restricts large scale applications such as in self-supervised learning setups.

This paper describes our submission to the DeftEval challenge Subtask 2. Our approach is based on a multi-task learning strategy, that jointly extracts concepts and relations between them, and achieves a score of 49.68 macro F1, which ranked our best system in place 27 of 51 participants. The major challenge turned out to be the strong label imbalance that made an evaluation using the official macro-averaged F1 score difficult.
The source code of our approach is publicly available.\footnote{\url{https://github.com/DFKI-NLP/defx}}

\section{DeftEval Challenge}

\begin{figure*}
    \centering
    \includegraphics[width=\linewidth]{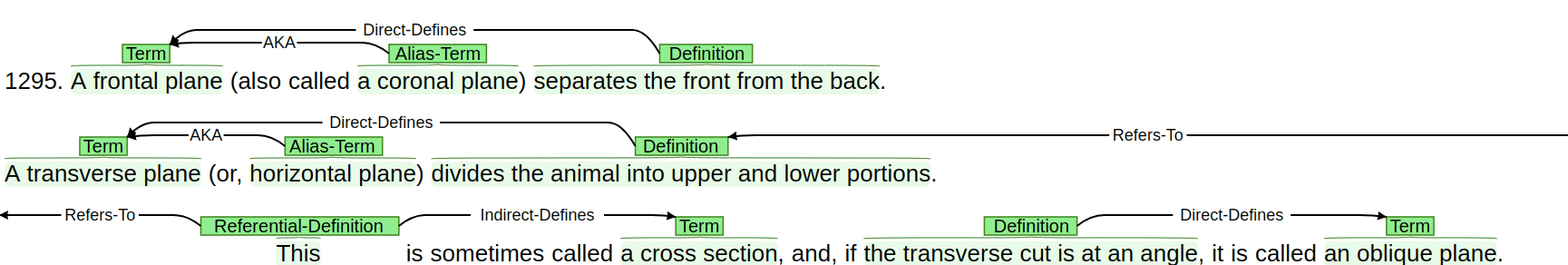}
    \caption{An exemplary input taken from the training set. It consists of three sentences and includes an \emph{Indirect-Defines} relation using a coreference relation that points to the definition.}
    \label{figure:annotated-example}
\end{figure*}

The DeftEval challenge\footnote{\url{https://competitions.codalab.org/competitions/22759}} \cite{shirani2020semeval} includes three subtasks: (1) classification of definition sentences, (2) sequence labeling of definition concepts and (3) relation extraction between concepts. The DEFT corpus \cite{deft-corpus} is used for evaluation, which consists of English textbooks scraped from an online-learning website.\footnote{\url{http://www.cnx.org}}
Compared to the previous manually annotated datasets WCL~\cite{wcl-corpus} and W00~\cite{w00-corpus}, the DEFT corpus is significantly larger and the examples are more diverse than simple is-a patterns. In contrast to WCL and W00 the DEFT corpus examples are not limited to definition sentences. Instead, they consist of windows of up to three sentences around a highlighted word in the source texts. Examples do not necessarily include a term-definition pair, but may include other concepts and relations, such as aliases or supplementary information. Additionally, definition relations do not have to be stated directly within one sentence, but can also be stated indirectly through a coreference relation to a preceding sentence. An example is depicted in Figure~\ref{figure:annotated-example}. The official metrics for Subtasks 2 and 3 are macro-averaged F1 scores evaluated on the labels in Table~\ref{table:evaluated_labels}.

\begin{table}
    \parbox[t]{.48\linewidth}{
    \begin{center}
        \begin{tabular}{l l}
            \toprule
            Subtask 2 & Subtask 3 \\
            \midrule
            Term & Direct-defines \\
            Alias-Term & Indirect-defines \\
            Referential-Term & Refers-to \\
            Definition & AKA \\
            Referential-Definition & Qualifies \\
            Qualifier & \\
            \bottomrule
        \end{tabular}
        \caption{Evaluated labels for Subtasks 2 and 3. \emph{Qualifies} is labeled as \emph{Supplements}.}
        \label{table:evaluated_labels}
    \end{center}
    }
    \hfill
    \parbox[t]{.48\linewidth}{
    \begin{center}
    \begin{tabular}{lrrr}
    \toprule
    Concept types          &  train &   dev &  test \\
    \midrule
    Term                   &   6385 &  1410 &   778 \\
    Definition             &   5943 &  1293 &   715 \\
    Alias-Term             &    671 &   164 &    86 \\
    Referential-Definition &    154 &    36 &    27 \\
    Qualifier              &    145 &    33 &    25 \\
    Referential-Term       &    134 &    25 &    15 \\
    \bottomrule
    \end{tabular}
    \caption{Counts of the Subtask 2 relevant concept types in the created dataset splits.}
    \label{table:split_sizes}
    \end{center}
    }
\end{table}

\section{Model}

We regard the DE task as the joint task of Named Entity Recognition (NER) and Relation Extraction (RE), and propose a system that closely follows the joint learning approach of~\newcite{Bekoulis2018JointER}. The NER part of their model is based on a BiLSTM CRF Tagger architecture~\cite{lstm-crf-mahovy,lstm-crf-lample}. The RE part projects each encoded token into a head and a tail space and classifies relations between all combinations of heads and tails, allowing for each token to have multiple relation heads.

\paragraph{CRF Tagger}
In our model each input example is first passed through a BERT model. The BERT layers are averaged for each token using a vector of learned weights, resulting in a token embedding. After passing through BERT, this token embedding is concatenated with the auxiliary input features and passed through the $n$ BiLSTM layers to learn a task-specific contextualized encoding $h$. This is projected into logits using a fully-connected layer and a CRF layer is trained to extract Subtask 2 labels. This subcomponent forms the \emph{CRF Tagger} baseline and the \emph{Simple Tagger} baseline is the same model without the CRF layer.
\paragraph{Relation Extraction}
RE is used as an auxiliary objective. As the challenge provided files in the DEFT format, which is a tab-separated format with a single relation tail per token, we modeled the constraint that every token could point to at most one relation tail. For each token $t_i$, the goal is to predict an output index $c_i$. The special case $c_i = 0$ corresponds to the negative class label $r_0$, if no relation exists between $t_i$ and any of the potential relation tails $t_j$. In all other cases the index $c_i$ is decoded to the $k$-th relation $r_k$ between token $t_i$ and token $t_j$.
The RE subcomponent uses only the hidden representation $h$ as input.\footnote{\newcite{Bekoulis2018JointER} propose to embed the predictions of the CRF layer and feed these as additional input into the RE subsystem. Since the focus was to improve the Subtask 2 submission score we skipped this step in order to maximize backpropagation effect in the jointly used BiLSTM layer.}
The hidden representation $h$ is passed through two projection layers $U$ and $W$ to extract representations for head and tail, respectively. For each token combination, the sum of the representations is passed through a $relu$ activation function in order to create a tensor of feature activations:
\begin{equation}
    M_{ij} = relu(U h_{i} + W h_{j} + b).
\end{equation}
The class logits are computed as follows:
\begin{equation}
    Q_{iz} =
    \begin{cases}
        V_0 max_{j'}(M_{ij'}) & \text{if } z=0 \\
        V_k M_{ij} & \text{else,}
    \end{cases}
\end{equation}
where the index $z$ is mapped to $j = floor((z - 1) / K^+)$ and $k = (z - 1) \;\mathrm{mod}\; K^+ + 1$ with $K^+$ being the number of non-negative relation labels.
Predictions are obtained via softmax:
\begin{equation}
    P(\hat{c_i} | t) = softmax(Q_i).
\end{equation}
The joint model is then trained on the sum of the losses for the two subtasks. We evaluated differently weighted losses, but found an equally-weighted sum to work best with respect to the Subtask 2 performance.

\paragraph{Auxiliary Input Features}
Additional input information augments the raw token input. String labels such as part of speech (POS) tags and NER labels are converted into vectors using a randomly-initialized learned embedding matrix.
For tokens that participate in a coreference cluster a binary indicator variable is added. Another set of binary variables indicates matches of rule-based patterns where each pattern match results in an activated indicator variable. Every binary variable is represented as one dimension of the additional input.

\section{Experiments}
\label{sec:setup}
The challenge provided labeled files split into training and development, while keeping the final test set hidden from the participants. We treated the official development set as our test set and created a new development set by randomly selecting 8 files from the training set. The resulting dataset sizes are listed in Table \ref{table:split_sizes}. There were several corpus changes shortly before and after the evaluation window of Subtask 2. We used the latest git commit before the evaluation period.\footnote{\url{https://github.com/adobe-research/deft_corpus/tree/ab1fb8951d0950a177e96}}

In a second step we converted the tab-separated DEFT files into a jsonl format and extended the data with additional information. Spacy v2.1.8 was used with the large English model\footnote{Spacy is available at \url{https://spacy.io/} and the used model is \emph{en\_core\_web\_large}} for POS tagging and rule-based patterns. In the case of coreference relations, these patterns targeted sentences starting with demonstrative determiners, e.g.\ ``this'' or ``these'', which are commonly used to refer to concepts in a previous sentence. We predicted coreference clusters using the implementation in AllenNLP v0.9.0 \cite{allennlp} proposed by \newcite{coref}.

Hyperparameter tuning was performed on the dev set using the Allentune implementation \cite{allentune}. The source code includes config files for the conducted experiments.

\section{Results}

Table~\ref{table:submission_scores} lists the results of the submitted runs. Using a majority-vote ensemble improved the score of the submitted joint model by ${\sim} 1$ point F1 to $49.68$ F1, ranking our submission in the 27-th place of 51 participants. A larger mix of models in an ensemble did not improve the performance over the standalone joint model. Precision and recall are balanced for the single-run model. Precision improved whereas recall degraded when multiple runs were evaluated in an ensemble.
\begin{table}
    \centering
    \begin{tabular}{lccc}
    \toprule
    Model & P & R & F1 \\
    \midrule
    Joint model {+}POS {+}coref ensemble & 52.05 & 48.04 & 49.68 \\
    Mix of ensembles & 51.74 & 46.52 & 48.74 \\
    Joint model {+}POS {+}coref & 49.32 & 49.56 & 48.56 \\
    \bottomrule
    \end{tabular}
    \caption{Official evaluation results of our Subtask 2 submissions. For the joint model both a single run and a majority vote ensemble of 10 repeated runs were evaluated. Additionally, a mixture of 6 model variations, each including 10 repeated runs, was combined into a large ensemble.}
    \label{table:submission_scores}
\end{table}

Table \ref{table:dev_results} shows the results of an ablation study over a subset of the tested model combinations. Each of the models was evaluated on both datasets using macro-averaged F1 score with 10 repetitions. Due to the strong label imbalance and small sample sizes all results show a high standard deviation. On the dev set the best F1 score was achieved by our submitted model. Best precision was achieved using a joint model and best recall using the CRF tagger. All of the joint models provide a higher precision than the baseline methods, while in recall they are worse than the CRF tagger but better than the Simple tagger. The F1 score is slightly higher for all models that employ a CRF layer. On the test set the CRF tagger performs best in terms of all metrics. The performance of joint models is slightly worse than the CRF tagger but better than the Simple tagger. Overall no single system is significantly better than the rest.

\begin{table*}
    \centering
    \begin{tabular}{llccc}
    \toprule
    Model & Split & P & R & F1 \\
    \midrule
    Simple tagger & dev & 55.62 +- 3.93 & 53.81 +- 2.87 & 53.80 +- 2.71 \\
    CRF tagger & dev & 56.05 +- 4.77 & \textbf{58.04 +- 5.91} & 55.18 +- 1.59 \\
    Joint model & dev & 58.09 +- 5.66 & 55.74 +- 4.47 & 55.04 +- 1.93 \\
    Joint model 
    {+}POS & dev & \textbf{58.16 +- 5.66} & 55.76 +- 4.45 & 55.08 +- 1.92 \\
    Joint model 
    {+}POS 
    {+}coref & dev & 57.64 +- 4.36 & 57.70 +- 6.34 & \textbf{56.27 +- 2.42} \\
    Joint model 
    {+}POS 
    {+}coref 
    {+}rules & dev & 57.19 +- 3.85 & 55.18 +- 4.00 & 54.85 +- 2.43 \\
    \midrule
    Simple tagger & test & 45.29 +- \hphantom{0}6.01 & 39.82 +- 5.77 & 40.29 +- 4.32 \\
    CRF tagger & test & \textbf{52.26 +- 12.74} & \textbf{50.06 +- 7.38} & \textbf{47.68 +- 7.97} \\
    Joint model & test & 45.43 +- \hphantom{0}9.56 & 42.52 +- 6.42 & 41.72 +- 6.26 \\
    Joint model 
    {+}POS & test & 45.44 +- \hphantom{0}9.49 & 42.61 +- 6.46 & 41.77 +- 6.20 \\
    Joint model 
    {+}POS 
    {+}coref & test & 50.73 +- 11.63 & 48.13 +- 6.24 & 44.65 +- 5.50 \\
    Joint model 
    {+}POS 
    {+}coref 
    {+}rules & test & 45.57 +- \hphantom{0}6.95 & 43.85 +- 6.72 & 42.63 +- 5.01 \\
    \bottomrule
    \end{tabular}
    \caption{Model ablations on the two splits with macro-averaged metrics on 10 repeated runs.}
    \label{table:dev_results}
\end{table*}

An analysis of the confusion matrices revealed that the most common misclassification is a confusion of \emph{Term} and \emph{Alias-Term}. A manual inspection of the training examples showed that many of these examples are very complex, and some examples were annotated incorrectly. A strong data bias towards pairs of \emph{Term} and \emph{Definition} resulted in a high rate of false positives for these types. Simple over- or undersampling did not help to combat these errors in our experiments.

We also observed that some possible annotations were missing. This was most notable for sentences starting with a demonstrative determiner, where the previous sentence included a definition concept. Furthermore, the annotation schema does not allow nested annotations although we have observed several examples where this would have been required.\footnote{
These issues might have been resolved after the evaluation window of the corpus, but we did not consult a repeated analysis to confirm this. Some sentences were duplicated for cases where annotations would have included an overlap.
}

In retrospect, Subtask 3 was likely too trivial to provide enough valuable information to improve upon Subtask 2. Alternative improvements could have been data augmentation for the minority class examples, and an improved use of the coreference and rule-based preprocessing.

\section{Conclusion}
In this paper we described our system for the joint extraction of definition concepts and relations among them, and reported results on the DEFT corpus, a definition extraction dataset of English textbooks. Despite the marginal improvements of our joint approach we also provide a robust setup of baseline methods, that we believe to be helpful to the community for experimentation and subsequent application in downstream tasks.

\section*{Acknowledgments}
This work has been supported by the German Federal Ministry of Education and Research as part of the projects DEEPLEE (01IW17001) and BBDC2 (01IS18025E), and by the German Federal Ministry of Economics Affairs and Energy as part of the project PLASS (01MD19003E).

\bibliographystyle{coling}
\bibliography{defteval}

\end{document}